\documentclass[letterpaper, 10 pt, conference]{ieeeconf}  
\usepackage{cite}
\usepackage{amsmath,amssymb,amsfonts}
\usepackage{algorithmic}
\usepackage{graphicx}
\usepackage{textcomp}
\usepackage{xcolor}
\usepackage{hyperref}
\usepackage{wrapfig,lipsum,booktabs}

\IEEEoverridecommandlockouts                              

\title{\LARGE \bf
Estimating Heterogeneous Causal Effect of Polysubstance Usage on Drug Overdose from Large-Scale Electronic Health Record
}

\author{Vaishali Mahipal$^1$, Mohammad Arif Ul Alam$^{1,2}$\\
\textit{$_1$Department of Computer Science, University of Massachusetts Lowell, USA} \\
\textit{$_2$Department of Medicine, University of Massachusetts Chan Medical School, USA} \\
vaishali\_mahipal@student.uml.edu, mohammadariful\_alam@uml.edu
}

\begin{document}

\maketitle
\thispagestyle{empty}
\pagestyle{empty}

\begin{abstract}

Drug overdose has become a public health crisis in the United States with devastating consequences. However, most of the drug overdose incidences are the consequence of recitative polysubstance usage over a defined period of time which can be happened by either the intentional usage of required drug with other drugs or by accident. Thus, predicting the effects of polysubstance usage is extremely important for clinicians to decide which combination of drugs should be prescribed. 
Recent advancement of structural causal models can provide ample insights of causal effects from observational data via identifiable causal directed graphs. In this paper, we propose a system to estimate heterogeneous concurrent drug usage effects on overdose estimation, that consists of efficient co-variate selection, sub-group selection and heterogeneous causal effect estimation. We apply our framework to answer a critical question, `can concurrent usage of benzodiazepines and opioids have heterogeneous causal effects on the opioid overdose epidemic?' Using Truven MarketScan claim data collected from 2001 to 2013 have shown significant promise of our proposed framework's efficacy. Latest paper and codes can be found here \url{https://arxiv.org/abs/2105.07224}
\end{abstract}

\section{Introduction}
The precision medicine initiative \cite{5} launched by then president of the United States (Barack Obama) in the year 2015 brought the attention of the research community to several unaddressed problems in the domain of personalized medicine. Patients in real-world cohorts exhibit complex heterogeneity \cite{10} making it difficult for doctors to prescribe an effective treatment plan based on inferences from a homogeneous subgroup. In addition, the cohorts from which these individual level inferences are to be made have been growing enormously over the past few years. In the era of big data, it has become feasible to learn complex high-dimensional models from such healthcare data using tools such as deep learning. However, whereas the machine learning community has made great progress toward using such rich models for supervised prediction, personalized or precision medicine requires causal models, for which there is significantly less theoretical and practical guidance available.

Understanding the cause and effect relationship for a specific phenomenon on any given individual group is an extremely challenging task due to the heterogeneity associated with them \cite{chirag20}. Recently, subgroup discovery for a specific phenomenon has drawn attention to the statisticians who addressed it by developing factual and counterfactual outcome models and then applying regression analysis on the difference of the two using another classifier (say decision tree) \cite{su09}. For example, \cite{27} proposed the subgroup identification based on differential effect search algorithms. On the other hand, qualitative interaction trees (QUINT) method \cite{elise14}, the virtual twins (VT) method \cite{12} and causal rule sets for discovering subgroups with enhanced treatment effect methods \cite{48} are also investigated previously. Nagpal et. al. proposed mixture of experts approach with soft assignment to groups that allows greater expressiveness and can be optimized via gradient methods \cite{chirag20}. While subgroup identification is an important step towards heterogeneous treatment effect model, we still need a heterogenous causal effect model in the next step. We utilize Nagpal et. al. proposed subgroup discovery and followed by a structured causal model.

Evidence suggested that polysubstance usage has adverse effects on drug overdose, such as concurrent usage of benzodiazepines and opioids over time has been associated with opioid overdose \cite{sun17}. However, what are the causal effects of polysubstance usage (benzodiazepines and opioids) on drug overdose requires appropriate causal effect estimation strategy. In this paper, our contributions are two folds:
\begin{itemize}
\item We design a systematic structure of polysubstance usage heterogeneous causal effect estimation technique to find the effect of drug overdose by following appropriate preprocessing, heterogeneous subgroup discovery, structural causal model, causal graph generation, finding confounders and finally causal estimation models.
\item Applied our model to identify heterogeneous causal effects of concurrent usage of benzodiazepines and opioids on opioid overdose epidemic using IBM Truven MarketScan healthcare claim data.
\end{itemize}

\begin{figure*}
  \centering
  \includegraphics[width=0.65\linewidth]{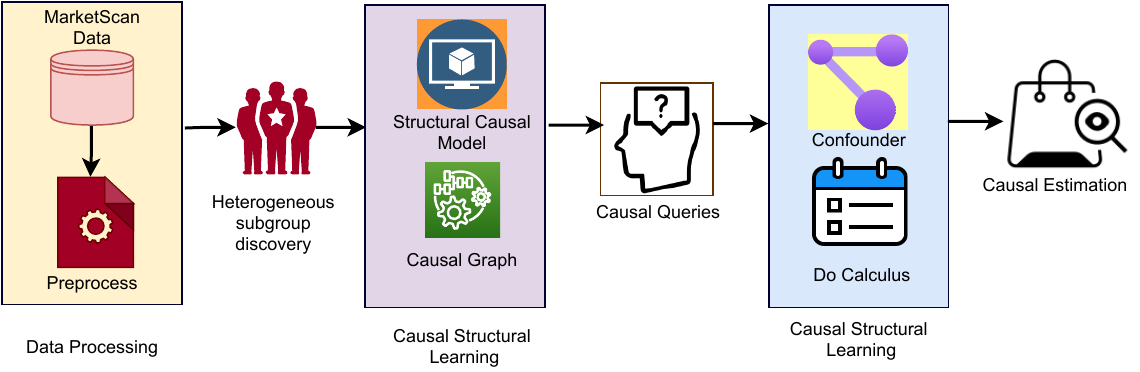}
  \vspace{-.1in}
  \caption{Overview of our proposed framework to identify heterogeneous causal effects of polysubstance usage on drug overdose}
  \label{fig:overview}
  \vspace{-.2in}
\end{figure*}

Fig. \ref{fig:overview} shows the schematic diagram of the different components of our proposed framework.

\section{Data Processing}
\subsection{Data and Outcomes}
We used de-identified Marketscan (Truven Health Analytics, Ann Arbor, MI) health claim data that provides patient level data on use and expenditures for the care of patients enrolled in private insurance plans. While initial sample size was 5,95,410 patients (1 January 2001 to 31 December 2013), we applied the following restricted steps to re-sample the data (i) excluded history of cancer patients (n=28,780), (ii) excluded age $<18$ and $>64$ (n=4,23,841 ), (iii) included at least one prescription of opioid (n=3,15,428), and, (iv) at least 3 years of continuous data (n=58994) which is our final cohort. Our primary outcome was an emergency room visit or inpatient admission for opioid overdose within a given calendar year, where we defined opioid overdose to be an admission ICD-9 code indicating either opioid related poisoning or a potential opioid related adverse event (such as respiratory depression) and an ICD-9 code corresponding to opioid overdose.

\subsection{Variables and Preprocessing}
Our key independent variable is the concurrent usage of benzodiazepine and opioids within a given calendar year. In this regard, we (i) identified opioid use by isolating all prescriptions for outpatient opioids, excluding prescriptions containing hydrocodone in a cough/cold formulation, (ii) isolated all prescriptions for a benzodiazepine and examined the degree of temporal overlap between prescriptions among individuals who filled a prescription for both classes of drugs and (iii) defined a similar interval for a benzodiazepine prescription and quantified the total number of opioid prescription days that overlapped with a benzodiazepine prescription days. Our central assumption is the concurrent use which we defined as with at least one day of overlap in a given calendar year. Apart from the binary variable assumptions, we also  considered several controls for patients' demographics and health that include age and gender. We obtained the variables from our source claims data while ICD-9 codes were used to control for comorbidities that include diabetes, mellitus, congestive heart failure, and so on for one year and onward. Apart from that, we also controlled for total healthcare spending in the time period before the first opioid prescription in a given year. In this regard, we separated all pharmacy, inpatient and outpatient claims submitted before the first opioid prescription of a given year. We then added all of the spending across these claims and divided it by the number of calendar days in the interval between the first day of the year and the date of the first opioid prescription day.

\section{Subgroup Discovery}
We used a heterogeneous effect mixture model (HEMM) \cite{chirag20} for identifying subgroups with enhanced treatment effect. This method hypothesized that a latent variable determines the concurrent usage of drug effect of each individual. Moreover, individuals with similar characteristics belong to the same latent subgroup, resulting in similar responses to concurrent drug usage across the subgroup. In this model, a Bayesian network was used to model these subgroups, specifically as a mixture model, along with their corresponding concurrent drugs usage effects.

\section{Causal Structural Learning}
A Structural Causal Model (SCM) that represents the causal relationship among covariate variables, can be represented as, $M = \rangle U, V, F, P(u)\langle$ tuple, where $U$ (exogenous) is determined by factors outside the model, $V$ (exogenous) is determined by variables in the model, $F={f_1,f_2,..f_n}$ where $f_i$ assigns a value to the corresponding $V_i\in V$ i.e. $v_i \leftarrow f_i(pa_i,u_i)$ and $P(u)$ is a probability function defined over the domain of $U$. Each $M$ is associated with a graph, known as a graphical causal model, $G$, which consists of a set of nodes or vertices that represent the variables $U$ and $V$ , and a set of edges that represents the functions in $f$. $G$ contains a node for each variable in $M$ and a directed edge from $X$ to $Y$ if $X$ is in the function $f$ of $Y$ ($X$ is a direct cause of $Y$ ) in $M$ \cite{pearl09}. Causal structure learning algorithms are used to generate causal directed acyclic graphs (DAG) from observational data \cite{heinz18}. Given a data set of opioid users with covariates, the framework estimates a possible causal graph that could be generated given the distribution using structure learning algorithms (SLAs) \cite{raghu18}. These SLAs learn the underlying causal graph at different levels of granularity under different assumptions. Based on these assumptions, multiple causal graphs were estimated using SLAs. We have considered the following seven SLAs for this framework: 1) PC Algorithm, 2) RFCI, 3) GES, 4) GDS, 5) TABU Search, 6) MMHC, and 7) LINGHAM \cite{glover98}. We then used a majority voting scheme on the causal relationships estimated using the SLAs i.e., if majority votes from all algorithms ($m=7$) counts $\frac{m}{2}+1$ then we support $X\rightarrow Y$ as a causal relationship. More specifically, if $G$ has $n$ nodes generating $n^2$ directed causal edges, then any edge $x\rightarrow y$ will be part of $G$ if majority votes $V(x,y)$> $\frac{m}{2}+1$.

\section{Causal Identification}
With the graph $G$ encoded with the causal relationships, we can now ask causal questions to understand how variables influence each other. This allows us to run virtual experiments to estimate treatment effects. The probability that $Y = y$ when we intervene to make $X = x$ is denoted by, $P(Y = y|do(X = x))$. Using do-expressions and a causal graph, we can ask causal questions and untangle the causal relationships. For our causal questions, $X$ is the treatment variable and $Y$ is the outcome. We used backdoor adjustment derived using the rules of do-calculus to find a set of confounding variables, Z that must be adjusted depending on the causal query in the form of $P(Y|do(X))$ and the graphical model \cite{pearl09}. The model automatically finds the confounding variables, can distinguish between confounders and mediators, and provides the causal identification formula. The causal identification formula provides a mathematical transformation between the observational reality and the corresponding experimental reality. The resultant formula can be used to evaluate the causal effect.

\section{Estimation}
Once we have these three sets of variables, treatment $X$, confounders $Z$, and outcome $Y$, we can compute the conditional probabilities in the adjustment formula using the probability distribution to compute the interventional effect. in a sub-group. Computing each of the conditional probabilities in the adjustment formula is computationally expensive and may not lead to possible results with small sample sizes. Instead, we use machine learning algorithms to estimate the probability distribution from the data, that is we use treatment $X$ and confounders $Z$ as input to an ML algorithm to estimate the outcome $Y$ . For this step, we split our data into training and testing sets, each containing $X$, $Y$ , and $Z$. We train two ML models, one based on observational data and another based on experimental data, using the training dataset. We use the testing set to predict $Y$ from $X$ and $Z$ using the trained ML models. After that, we use bootstrapping to randomly subsample $X$ and predict $Y$ for $k$ iterations. This gives us an estimation of the effect of $X$ on $Y$ in each sub-group and used to estimate the mean and confidence interval. Our approach allows straightforward querying of the model with varying observations of covariates related to concurrent usage of benzodiazepines and opioids. For instance, a patient who is included as a concurrent user of benzodiazepines and opioids, the probability of high Negative Affect (NA) score when the user is identified as opioid overdosed and had lower Negative Affect (NA) score before opioid overdose has been reported.
\begin{equation}
Pr(Post NA = H | Overdose = Y, Pre NA = L)
\end{equation}

\section{Experimental Evaluation}

\begin{table}[!h]
\begin{scriptsize}

  \caption{Opioid-Denzodiazepines data Statistics}
  \label{tab:statistics}
  \centering
  \begin{tabular}{|p{5cm}|p{0.5cm}|p{2cm}|}
      \hline
    Total Covariate Dimension & 1,228 & 5 continuous and 1,223 binary \\  
 \hline
     ICD-9 Diagnostic Codes & 1,013 & Binary \\    
 \hline
     CTP Procedure Codes & 171 & Binary \\ 
 \hline
    Hand-Crafted Commodities & 41 & Binary \\  
 \hline
    Daily Morphine Equivalent, Total number of visits, Gender, Age, daily health spending (\$) & 5 & Continuous\\    
 \hline
 \\
 \hline
     {\bf  Opioid Overdose} & {\bf Y=1} & {\bf Y=0} \\  
 \hline
     Treated (Opioid+Benzo) (T=1) & 63 & 5,425\\  
 \hline
      Not Treated (Opioid only) (T=1) & 118 & 53,389 \\  
 \hline
       Total & 181 & 58,814 \\
 \hline
  \end{tabular}
  \vspace{-.3in}
\end{scriptsize}
\end{table}

\subsection{Experimental Setup}
We have chosen a total of 1,228 covariates in our model that consists of 5 continuous and 1,223 binary variables as described in Table ~\ref{tab:statistics}. Table ~\ref{tab:statistics} also stated descriptive statistics of treatment where we considered the usage of opioid and benzodiazepines concurrently as a treated population, where the number of opioid overdose incidences are 63 and 118 for treated and not treated use cases respectively. We used a subgroup discovery algorithm on this sample of dataset using the HEMM \cite{chirag20} method. In this method, at first, we assigned individuals to an enhanced effect subgroup of varying size matrix. Then we chose the subgroup $k$ with the largest main effect $\gamma_k$ (as stated in \cite{chirag20}) and varied threshold which has been applied to the corresponding membership probability ($p(Z=k|X)$) returned by the model. Here $p$ is the group membership probability, $k$ is the number of groups and $X$ is the covariate matrix. In the second step, we focused on estimating the effect of different treatment (opioid+benzodiazepines or opioid only) strategies in each subgroup by establishing the causal relationship between measured covariates as stated in Table \ref{tab:statistics}. Demographic data and computed mean were then used as features to estimate causal graphs using (SLAs) \cite{raghu18}. Table \ref{tab:features} shows the top features measured from our sample dataset. For each algorithm, we learned the causal relationship among the covariates and then estimated a causal graph, and finally generated a total of 7 causal graphs. Based on the SLAs, we computed the matrix $V$ for voting which resulted in a 1,228 x 1,228 matrix, where each row has a variable for which we have the vote count in the columns for all 1,228 variables. Finally, we compute the mean and confidence interval of all subgroups to estimate heterogeneous treatment effects across the entire population.

\subsection{State-of-Art Regression Analysis}
At first, we did state-of-art statistical analysis to find associations. We used multivariate logistic regression to estimate the association between the treatment (concurrent usage of benzodiazepine and opioid) and the outcome (opioid overdose) among opioid users \cite{sun17}. The dependent variable of the regression is emergency room visit due to opioid or not (1 or 0) in the given calendar year, while independent variable is whether the opioid user concurrently used benzodiazepine in the given year. We considered the population attributable fraction (PAF) of concurrent usage of benzodiazepine and opioid to estimate opioid overdose risks.
\subsection{State-of-Art Data for Evaluating Model Performance}
Since, there are no way to evaluate our proposed causal effect model on observational data, we utilize Infant Health and Development Program (IHDP) dataset, which is a popular data for evaluating heterogeneous treatment effects model that provides causal effect ground truths\cite{ihdp}. The data consists of 25 real covariates generated from a randomized experiment to evaluate the benefit of IHDP on IQ scores infants (3 years old). Then, some of the treated population has been removed under selection bias condition generating 608 control  and 139 treated patients (747 total). The outcomes and counterfactuals were simulated using the standard non-linear `Response Surface B'.

{
\begin{table}[!h]
\vspace{-.1in}

\caption{Mean absolute error of Average Treatment Effect (ATE) on IHDP data comparisons among state-of-art causal effect models with or without sub-group selection module}\label{tab:ihdp_results}
\begin{center}

\begin{tabular}{|c|c|c|}
\hline
Method & $\epsilon_{no_subgroup}$ & $\epsilon_{with_subgroup}$  \\
\hline
CFRW &.25$\pm$.01 & .23$\pm$.01\\ 
\hline
CEVAE &.34$\pm$.01 & .31$\pm$.01\\ 
\hline
Ours &.20$\pm$.01 & .18$\pm$.01\\  
\hline
\end{tabular}

\end{center}
\vspace{-.3in}
\end{table}
}
\subsection{Results}
{\bf IHDP Datasets}: To evaluate our model's efficacy, we consider IHDP data where intervention is whether an infant gets treatment or not, and the outcome is whether cognitive test scores (CTS) is higher after treatment or not. We use absolute error of average treatment effect (ATE), $\epsilon$, which is the absolute differences between ground truth and our model predicted ATE (smaller is better). Table \ref{tab:ihdp_results} shows details of the $\epsilon$ measures on few latest causal effect estimation models. Table \ref{tab:ihdp_results} shows that our method performs higher than CFRW and CEVAE causal effect models. Also we showed that subgroup selection model always increases the performance for any state-of-art causal effect models.

\begin{table}[!h]
\begin{scriptsize}
  \caption{Top Features of the enhanced effect subgroup $k$ discovered by HEMM \cite{chirag20} on Opioid dataset, where the ratio represents the increase in prevalence over the other group (k=2), i.e., ratio = 0.5 means no increase in prevalence over other group}
  \label{tab:features}
  \centering
  \begin{tabular}{|p{2.5cm}|p{5.2cm}|}
      \hline
    Musculoskeletal System & 1.0 spinal curve (kyphosis, lordosis, scoliosis), 1.0 ankle fracture, 1.0 sprains/strains of hand and wrist\\  
 \hline
     Nervous System & 1.0 extrapyramidal diseases/movmt. disorders, 1.0 idiopathic peripheral neuropathies, 1.0 headaches\\    
 \hline
    Integumentary System & 1.0 cellulitis and abscess of finger and toe,   1.0 local skin infections, 1.0 psoriasis and similar disorders\\ 
 \hline
    Reproductive System & 1.0 female infertility, .82 testicular dysfunction, .82 disorders of penis \\  
 \hline
    Circulatory System & 1.0 hypertensive heart disease, .79 other disorders of circulatory system,  .70 cardiac dysrhythmias\\    
 \hline
    Nutrition& 1.0 BMI, 1.0 b-complex deficiency,  .71 disorder of electrolyte/acid-base balance\\    
 \hline
    Psychology & 1.0 suspected mental health condition,  .58 adjustment reaction, .55 nondependent abuse of drugs\\    
 \hline
     Respiratory System & 1.0 other diseases of respiratory tract, .74 deviated nasal septum, .69 influenza \\    
 \hline
  \end{tabular}
\end{scriptsize}
\end{table}

{\bf MarketScan Dataset}: Statistical analysis on the sample dataset showed that, among opioid users, patients who used opioid and benzodiazepine concurrently were older (44.5 v 42.4, P<0.001) and most of them are women (35\% v 43\%, P<0.001) compared with those who used opioid only. Using the logistic regression model, we calculated the PAF for concurrent prescription of opioid and benzodiazepine to be 0.15 (95\% confidence interval 0.14 to 0.16) among all opioid users, suggesting that eliminating concurrent usage of benzodiazepine and opioid can reduce the population risk for an opioid overdose related inpatient admission or emergency room visit by 15\%. Using our proposed heterogeneous causal effect model, we performed the effect of opioid overdose experiments based on the causal graph and the selected cohort from the observational data where we considered concurrent usage of benzodiazepine and opioid as a treatment among opioid users and only usage of opioid as not treatment group. From our proposed framework, we estimated the causal expectation for benzodiazepine co-prescribing to be 0.29 (95\% confidence interval 0.25 to 0.30) among all opioid users, which is higher than the regression analysis provided. This analysis suggested that concurrent usage of benzodiazepine/opioid usage has a high causal effect (29\% higher impact) over opioid overdose.
\section{Conclusion}
It is well known to the experts that polysubstance usage is one of the major causes of drug overdose. However, there were no studies before that utilized causal models with efficient subgroup discovery-aided causal estimation techniques to estimate heterogeneous causal effects of polysubstance usage on drug overdose. In this study, we particularly investigated existing causal inference models and developed an efficient framework to identify the causal effects of concurrent usage of opioid and benzodiazepines on opioid overdose. Our generic framework can be a foundation of investigating concurrent events' causal effects on any outcome that involves heterogeneity.

\end{document}